\documentclass[a4paper,11pt]{article}

\usepackage{bibspacing}
\usepackage{verbatim}
\usepackage{setspace}
\usepackage{fourier}
\usepackage{color}
\usepackage{graphicx}
\usepackage{url}
\usepackage[affil-it]{authblk}
\usepackage{amsmath}
\usepackage{wrapfig}
\usepackage{array}
\usepackage[T1]{fontenc}
\usepackage{times}
\usepackage{booktabs}
\usepackage{indentfirst}
\usepackage[colorlinks,linkcolor=red,anchorcolor=blue,citecolor=green]{hyperref}

\begin{document}

\title{ \textbf{\textit{Rb-PaSta}Net}: A Few-Shot Human-Object Interaction Detection Based on Rules and Part States}

\author{Shenyu Zhang, Zichen Zhu, Qingquan Bao}
\affil{Shanghai Jiao Tong University}
\date{}
\maketitle
\thispagestyle{empty}

\begin{abstract}
Existing Human-Object Interaction (HOI) Detection approaches have achieved great progress on non-rare classes while rare HOI classes are still not well-detected. In this paper, we intend to apply human prior knowledge into the existing work. So we add human-labeled rules to \textbf{\textit{PaSta}Net} and propose \textbf{\textit{Rb-PaSta}Net} aimed at improving rare HOI classes detection. Our results show a certain improvement of the rare classes, while the non-rare classes and the overall improvement is more considerable. 
\end{abstract}

\section{Introduction}

When building an intelligent system, understanding human activities from still images plays a critical role. As a sub-task of visual relationship comprehension \cite{tin19}, Human-Object Interaction (HOI) infers types of interactions through retrieving human and object locations. Related to human and object understanding, HOI will boost activity understanding \cite{tin2}, imitation learning \cite{tin1}, etc. 
\par
\indent Generally, this high-level cognition task is addressed in one-stage \cite{pasta7}, i.e. directly mapping pixels to activity concepts. Closer to our work, Li \textit{et al.} \cite{1} takes advantages of \textit{part-level semantics} and builds a human activity knowledge engine to infer interactiveness. However, nearly all state-of-the-art methods encounter few-shot classes' performance bottleneck due to rare training datasets.
\par
\indent In light of this, we propose a \textbf{\textit{Rb-PaSta}Net} (Rule-based Part State Net) to augment the existing work \textit{PaSta}Net \cite{1}. With the introduction of human prior knowledge as rules, we believed \textbf{\textit{Rb-PaSta}Net} would have more information about the physical world besides training data. Specifically, we manually label the weights of each human body part in 162 less-than-ten-shot HOI classes. Two versions are introduced: one consists of weights in Decimal numbers derived from the average of three authors' labels, the other Boolean numbers derived from the former version. These weights are added to part attentions, which means the importance of certain body parts in an HOI class, to strengthen the learning of few-shot HOI classes.

\par
\indent Finally, our method achieves some insignificant improvement on few-shot HOI classes, e.g. the Decimal version makes 0.13 mAP improvement of the rare classes on HICO-DET \cite{tin3}. Meanwhile, we gladly found that the Boolean version has improved the non-rare classes and the overall mAP by 0.22 and 0.2. So after we point out the deficiencies of current human-based rules, we propose some viable approaches to enlarge the improvement.

\section{Related Work}
Our work is in the field of Human-Object Interaction (HOI) where image-based, instance-based and body-part-based patterns are mainly used.

\paragraph{Human-object Interaction}
Most of the daily human activities involve HOI \cite{tin3,li2020detailed} . Thanks to Deep Neural Networks (DNNs), many great improvements have been made in the detection of such events \cite{tin3,tin9,tin21}. Chao \textit{et al.} \cite{tin3} combined visual features and spatial locations to construct a multi-stream model. Qi \textit{et al.} \cite{tin21} proposed Graph Parsing Neural Network (GPNN) incorporating DNN and graphical model to iteratively update states and classify pairs. Gao \textit{et al.} \cite{tin9} developed an instance centric attention module to increase the information from the region of interest and improve the HOI classification. Li \textit{et al.} \cite{tin} explored interactiveness knowledge learned from various HOI datasets, implicitly increasing the training data for rare HOI classes. While these works have contributed to the improvement of HOI detection, the progress of the few-shot HOI classes is insufficient \cite{1}.

%%%need considering how to revise...
%comments: 
%Page 2, first paragraph: Ends with the phrase "While these works have contributed to the improvement of HOI detection, the progress of the few-shot HOI classes is insufficient."   ----   however, no evidence is given to support this statement. Why is progress insufficient?
%他咋不问我intro
%another comment:
%Page 2, second paragraph: "Despite that, the mAP of one-shot set in the current state-of-the-art is still below 0.3"   ----   This needs a reference to back up the statement.

%其实就一个问题，pasta是SOTA了，第一段进步不足当然指的就是目前这个sota还不足，第二段sota below 0.3 其实也指的是pasta，怎么让他看懂。。。
%%%

\paragraph{Part States}
Lu \textit{et al.} \cite{pasta38} proposed a discrete set of part states through tokenizing the semantic space and bases a sort of basic descriptors on segmentation \cite{pasta15}. Furthermore, Li \textit{et al.} \cite{1} utilized states of 10 human body natural parts to represent activities and reasons out the activities with \textit{part-level sementics}. \textit{PaSta}Net makes great improvements and reaches the-state-of-the-art in both full and few-shot tasks. Despite that, the mAP of one-shot set in the \textit{PaSta}Net is still below 0.3. To improve that, in this paper, we mainly focus on few-shot problems in HOI.

\begin{figure}[!h]
\vspace{-10.5pt}
\begin{center}
\includegraphics[width=.9\textwidth]{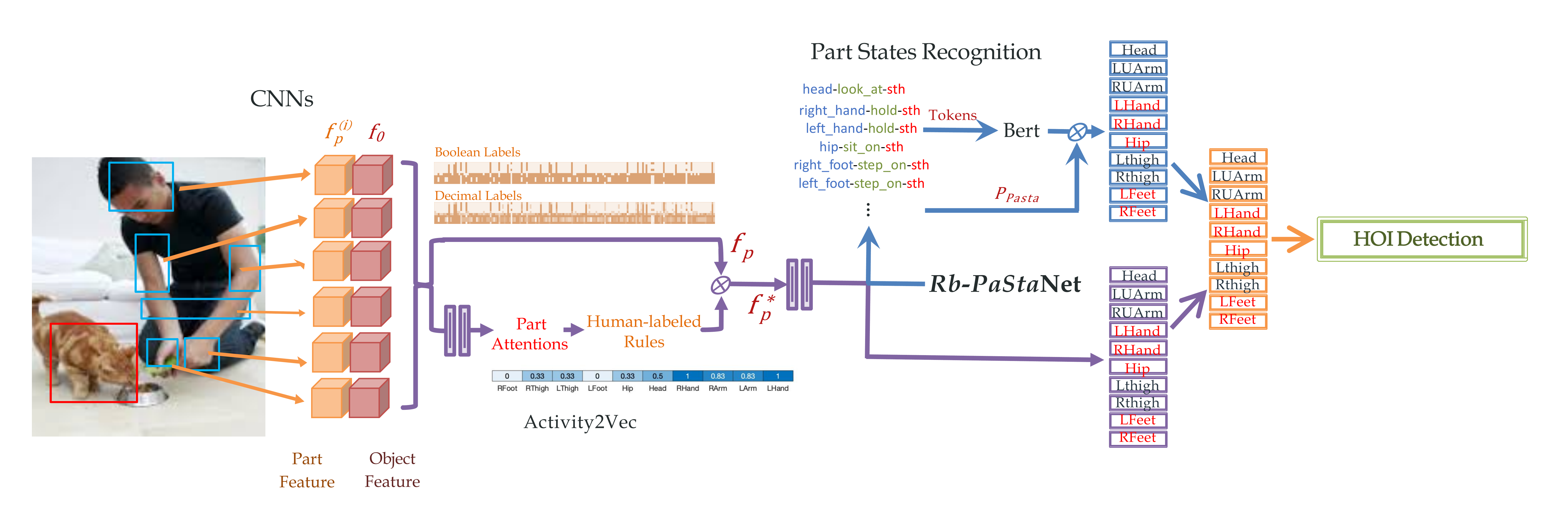}
\end{center}
\vspace{-27pt}
  \caption{Overview of \textbf{\textit{Rb-PaSta}Net}}
  \vspace{-20pt}
  \label{figure0}
\end{figure}

\section{Approach}

In this section we introduce the construction of \textbf{(\textit{Rb-PaSta}Net)} in Figure\ref{figure0} to tackle the few-shot problems. Why current DNN-based models do not perform well in few-shot HOI classes may lie in insufficient information offered by training datasets. Therefore, our method introduces human prior knowledge to reinforce learning in rare HOI classes. Considering \textit{PaSta}Net \cite{1} using body parts' action as a medium to infer HOI, adding prior rules into each body parts action weights can be viable. In the following paragraphs, the choice of the dataset, the method to label and the process of rules construction are specified.

\paragraph{Data} 
We conduct model training based on the \textit{PaSta}Net database, with more than 200k training examples. The existing network divides object-action into 600 classes(including the no-interaction ones). A total of 162 classes only appear in the training set less than ten times. Our experiments are targeted at optimizing existing networks in terms of these rare classes.

\begin{figure}[!h]
\begin{center}
\includegraphics[width=.9\textwidth]{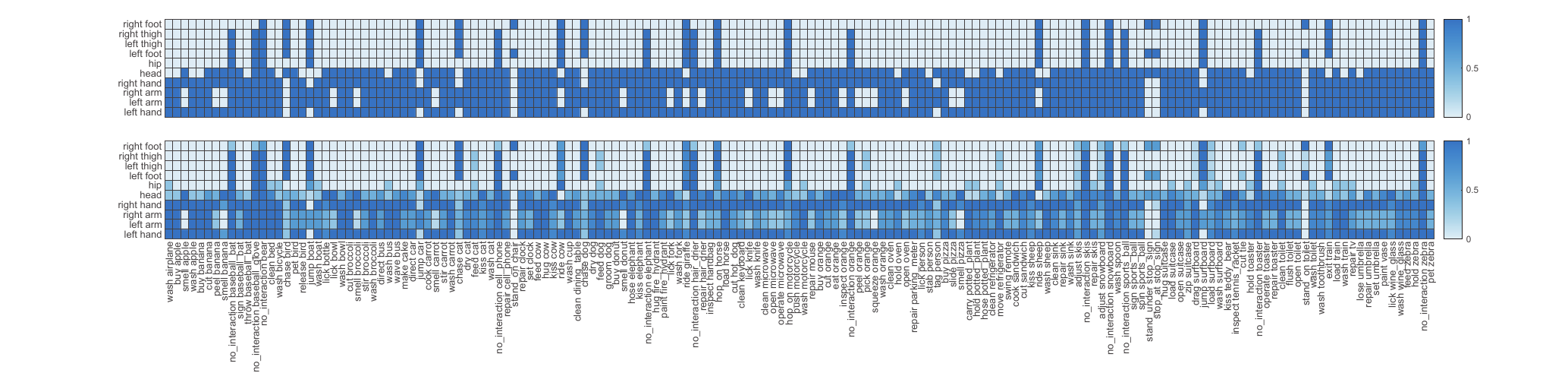}
\end{center}
\vspace{-25pt}
  \caption{the Boolean and Decimal label matrix}
  \label{hot}
  \vspace{-20pt}
  
\end{figure}

\paragraph{Label} 
As for the rare classes, three authors perform manual annotations respectively based on the body parts' involvement in the corresponding object-action according to our prior knowledge. In our annotation, label 1 indicates a strong correlation; 0.5 indicates a certain degree of correlation; 0 indicates irrelevance. As for common classes, all parts are labelled 1. Then we average three annotations as one type of label. The average value is distributed between 0 and 1. Besides, to find a better rule, we map the resulting Decimal label to the Boolean label (True if the decimal label is no less than 0.5, or False otherwise). The two groups of labels and the All-True control group (original label) were trained separately. Table\ref{sheep} shows an example of labelled weights in an HOI "feed a cat" and all weights are displayed in Figure\ref{hot} where the upper one represents the Boolean version and the lower one represents the Decimal version.

%- At the end of Page 2, it is not clear why the thresholding was carried out and what the benefit of this would be. 【 他可能问的是为什么要有bool version？】
%- In Table 1, the decimal values for Head, RHand, RArm and LHand do not correspond to the Boolean values.  【反正骑羊要换的

\paragraph{Implementation}
In \textit{PaSta}Net \cite{1}, all features will be initially input to a \textbf{Part Relevance Predictor} telling a body part's importance in an action. Formally, a certain attention is
\begin{equation}
    a_i = \mathcal{P}_{pa}(f_p^{(i)},f_o)   
\end{equation}
where $\mathcal{P}_{pa}(\cdot)$ is the part attention predictor and $f_p^{(i)}$, $f_o$ indicate features of a part and an interacted object respectively. In \textbf{\textit{Rb-PaSta}Net}, we introduce rules :
\begin{equation}
    a_i^{Rb} = a_i \cdot a_{rules}
\end{equation}
where $a_i^{Rb}$ represents attentions added with rules and $a_{rules}$ indicates weights we have labeled in the last paragraph. Then we compute scores and cross-entropy loss like \textit{PaSta}Net with $a_i^{Rb}$ instead of $a_i$.

\begin{table}[!h]
\centering
\begin{tabular}{cccccccccccc}
\toprule
\multicolumn{7}{r}{Body Parts} \\
\cmidrule(r){2-11}
Method & RFoot	& RThigh &	LThigh & LFoot &	Hip	& Head &	RHand	& RArm &	LArm &	LHand &Average \\
\midrule
Original & $1$ & $1$ & $1$ & $1$ & $1$ & $1$ & $1$ & $1$ & $1$ & $1$ & $1$\\
Decimal & $0$ & $0.33$ & $0.33$ & $0$ & $0.33$ & $0.5$ & $1$ & $0.83$ & $0.83$ & $1$ & $0.52$\\
Bool & $0$ & $0$ & $0$ & $0$ & $0$ & $1$ & $1$ & $1$ & $1$ & $1$ & $0.5$\\
\bottomrule

\end{tabular}

\vspace{-7pt}
\caption{The weights of "feed a cat"}
\label{sheep}
\end{table}
\vspace{-15pt}

\section{Experiment}

\paragraph{Settings} 
We adopt one HOI datasets HICO-DET \cite{tin3} with 600 HOI categories on 80 objects categories and 117 verbs. We first use best pre-trained Activity2Vec with \textit{instance-level} \textit{Pasta} labels \cite{1} and then fine-tune \textbf{\textit{Rb-PaSta}Net} on HICO-DET. All testing data are separated from pre-training and fine-tuning. We follow the metrics of \cite{tin3} for results and PaSta detection. The fine-tuning takes 2M iterations and the learning rate is 1e-3 with 1:4 of positive and negative samples. A late fusion strategy is adopted.

\begin{table}[h!]
\centering
\begin{tabular}{lllllll}
\toprule
Method               & Full(def) & Rare(def) & Non-rare(def) & Full(ko) & Rare(ko) & Non-rare(ko) \\
\midrule
PaStaNet             & 21.92     & 20.44     & 22.37         & 23.86    & 22.31    & 24.33        \\
Rb-PaStaNet(Boolean) & 22.12     & 20.54     & 22.59         & 24.04    & 22.43    & 24.52        \\
Rb-PaStaNet(Decimal) & 21.94     & 20.57     & 22.35         & 23.86    & 22.46    & 24.28       \\
\bottomrule

\end{tabular}
\vspace{-0pt}
\caption{Results on HICO-DET. We follow the evaluation metrics in \cite{tin3}: def means \textit{Default} setting where the full test set is detected, while ko means \textit{Known Object} setting where the target object category is given.}
\vspace{-15pt}
\label{result}
\end{table}

\paragraph{Results}
As Table\ref{result} shows, Boolean and Decimal versions of \textbf{\textit{Rb-PaSta}Net} achieve 0.1 and 0.13 mAP improvements on rare HOI classes. Although the human labels are not very precise concerning they are based on few people's intuition and comprehension, the result has proved that by applying human prior knowledge the mAP can be improved. Meanwhile, the result shows that the Boolean version has improved the non-rare classes and the overall mAP by 0.22 and 0.2. After analysing the scores of each classes, we find out that the rules are also influencing those non-rare classes(just think those classes share the same label-[1,1,1,1,1,1,1,1,1,1]). So we have proposed a few possible approaches that may increase the competitiveness of \textbf{\textit{Rb-PaSta}Net}:
\begin{itemize}
    \item label all the 600 HOI classes more comprehensively and rigorously
    \item label few-shot pictures instead of considering the rare classes as a whole
\end{itemize}
%In other words, our prior knowledge is not precisely quantified with mere 0, 0.5 and 1. Besides, we are not able to distinguish whether a body part is indeed involved in a certain activity. That leads to excessive 1 label and undermines our rules' effectiveness. Therefore, we think human experience fails to be quantified and added to machines' weights. For further study, we intend to transfer weights machines learned from non-rare HOI classes to rare classes since we find HOI classes with the same verb tend to have similar body part attention when labelling.

\section{Conclusion}
%In this article, based on the existing \textbf{\textit{PaSta}Net} , we propose \textbf{\textit{Rb-PaSta}Net} instead and demonstrate that simple human labels hardly work. We added manually labelled action-part rules, which aim to improve the recognition accuracy of rare samples by adjusting the training weights of different parts. The Boolean and Decimal labels are compared with the original network model results. The detection accuracy of the rare class of both versions has been slightly improved by 0.1 and 0.13 mAP. We figure out transferring learning from machines themselves may be the more promising approach.
In this article, based on the existing \textit{PaSta}Net, we propose \textbf{\textit{Rb-PaSta}Net} instead. Our goal is to improve on rare classes by adjusting the training weights of different parts. The experiment result has proved our method is feasible, but it also leaves much room for improvement. The probable reason is that our rule itself is not accurate enough because it is generated by three authors. In the future, we hope to get a more accurate and detailed version with the help of volunteers. We believe a better rule can make more improvement.

\section*{Acknowledgments}
We would like to express our very great appreciation to Cewu Lu and Yong-Lu Li for their constructive suggestions and guidance throughout this research work. We would also like to extend our thanks to Liang Xu for his assistance in terms of codes. Their patience and carefulness have been very much appreciated.

\bibliography{main}

\end{document}